\pdfoutput=1

\documentclass[11pt]{article}

\usepackage[final]{acl}

\usepackage{times}
\usepackage{latexsym}

\usepackage[T1]{fontenc}

\usepackage[utf8]{inputenc}

\usepackage{microtype}

\usepackage{inconsolata}

\usepackage{graphicx}

\usepackage{times}
\usepackage{soul}
\usepackage{url}
\usepackage{amsmath}
\usepackage{amsthm}
\usepackage{booktabs}
\usepackage{algorithm}
\usepackage{algorithmic}
\usepackage[switch]{lineno}
\usepackage{multirow}
\usepackage{array}
\usepackage{csquotes}
\usepackage{xcolor}
\newcommand{\ignore}[1]{}
\usepackage{adjustbox}

\title{CRAB: A Benchmark for Evaluating Curation of Retrieval-Augmented LLMs in Biomedicine}

\author{Hanmeng Zhong \\
  \small{PatSnap Co., LTD.} \\
  {\tt \small{zhonghanmeng@patsnap.com}} \\\And
Linqing Chen\thanks{Corresponding author.} \\
  \small{PatSnap Co., LTD. } \\
  {\tt \small{linqingchen6@126.com}} \\\And
   Wentao Wu \\
  \small{PatSnap Co., LTD.} \\
  {\tt \small{wuwentao@patsnap.com}}
   \\\And
  Weilei Wang \\
  \small{PatSnap Co., LTD. } \\
  {\tt \small{wangweilei@patsnap.com}}
 \\}

\begin{document}
\maketitle
\begin{abstract}
Recent development in Retrieval-Augmented Large Language Models (LLMs) have shown great promise in biomedical applications. However, a critical gap persists in reliably evaluating their curation ability—the process by which models select and integrate relevant references while filtering out noise. To address this, we introduce the benchmark for \textbf{C}uration of \textbf{R}etrieval-\textbf{A}ugmented LLMs in \textbf{B}iomedicine (\textbf{CRAB}), the first multilingual benchmark tailored for evaluating the biomedical curation of retrieval-augmented LLMs, available in English, French, German and Chinese. By incorporating a novel citation-based evaluation metric, CRAB quantifies the curation performance of retrieval-augmented LLMs in biomedicine. Experimental results reveal significant discrepancies in the curation performance of mainstream LLMs, underscoring the urgent need to improve it in the domain of biomedicine. Our dataset is available at \href{https://huggingface.co/datasets/zhm0/CRAB}{https://huggingface.co/datasets/zhm0/CRAB}.
\end{abstract}

\section{Introduction}

Large Language Models (LLMs) have become powerful tools in general and specific domains \citep{openai2024,llama3,claude3,gemini1.5,chen2025streamlining}, due to their generative capabilities \citep{bang-etal-2023-multitask,guo2023closechatgpthumanexperts}. Despite their excellent performance \citep{largelanguagemodelsencode,palm2,openai2024,llama3,gpt4medicalchallenge}, LLMs still suffer from factual hallucinations \citep{factual,raunak-etal-2021-curious,Ji-2023}, caused by outdated knowledge \citep{rethinkingretrievalfaithfullarge} and limited domain-specific expertise \citep{chatgptgpt4generalpurposesolvers}.
Retrieval-Augmented Generation (RAG) addresses these shortcomings by retrieving up-to-date information from trusted sources. This approach has been shown to effectively mitigate hallucinations and knowledge gaps in various domains \citep{guu2020realmretrievalaugmentedlanguagemodel,NEURIPS2020_6b493230,pmlr-v162-borgeaud22a,JMLR:v24:23-0037}. In the biomedical domain, the incorporation of external knowledge not only enhances the existing capabilities of LLMs, but also provides the up-to-date information that they lack to accurately answer the biomedical queries \citep{biom1,biom2}.

\begin{figure}[t]
    \centering
    \includegraphics[width=\linewidth]{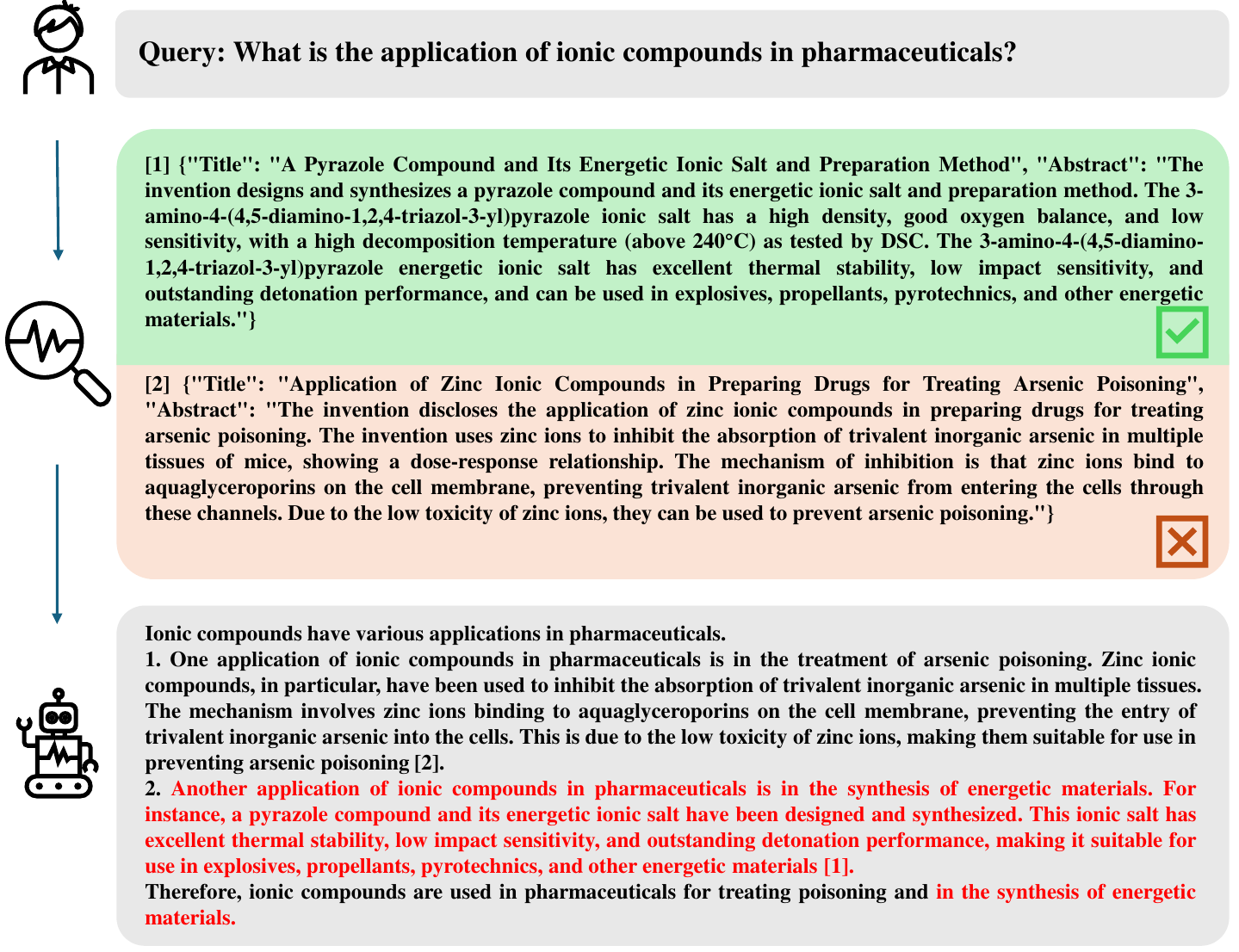}
    \caption{An example of weak curation ability. Here green and check marks indicate relevant, while red and cross marks indicate irrelevant. Red text represents content generated based on irrelevant references.}
    \label{noise example}
\end{figure}

While RAG improves performance, it also introduces challenges \citep{menick2022teachinglanguagemodelssupport,li-etal-2023-large}. If irrelevant content is retrieved, LLMs may generate invalid content in responses. For example, in Figure \ref{noise example}, given the query ``What is the application of ionic compounds in pharmaceuticals?'' and two references (one relevant, one irrelevant), the LLM generates irrelevant content about ``ionic compounds in energy materials''.
Despite advances in retrieval systems, irrelevant content remains a challenge. It generally falls into three categories: (1). partially relevant but can not answer the query; (2). completely irrelevant;  (3). factually incorrect. Currently, the difficulties regarding retrieval-augmented LLMs mainly focus on the first category. And distinguishing them from the relevant references is a prerequisite for correctly answering the query, which also requires comprehensive comprehension ability.
In the biomedical domain, precise citations in the responses are essential. Accurate citations enable independent verification and reproducibility, while robust curation ensures the scalability of responses. This is of great importance both in the research and application domains of biomedicine.\ignore{This level of rigor is critical, as even minor errors may lead to data misinterpretation and potentially compromise patient safety and treatment efficacy.} However, the existing retrieval-augmented benchmarks \citep{mirage,comprehensivepracticalevaluationretrievalaugmented} in the domain of biomedicine are no different from those in the general domain, focusing on evaluating the correctness of the answers and overlook the importance of curation.

In this paper, we introduce \textbf{CRAB}, the multilingual benchmark for \textbf{C}uration of the \textbf{R}etrieval-\textbf{A}ugmented LLMs in \textbf{B}iomedicine, evaluating the curation ability of retrieval-augmented LLMs to distinguish and utilize references when answering biomedical queries, avaliable in English, French, German and Chinese. In detail, CRAB focuses on the open-ended queries, addressing two key issues of curation evaluation based on the closed-ended queries: (1). predefined answers: LLMs may be able to answer the query without augmented references; (2). inflexibility in updating evidence: biomedical research is continually evolving, and predefined answers may quickly become outdated.
Moreover, open-ended queries in retrieval-augmented scenarios do not have standard answers, which is more conducive to focusing on the evaluation of curation.\ignore{the ground-truth answers of the open-ended queries can vary on the different augmented references: (a). the quality of the answer depends on the quality of the relevant references; (b). different LLMs may interpret the same reference differently. Therefore, CRAB can more focus on the curation evaluation.} For the augmented references, we apply LlamaIndex\footnote{\url{https://github.com/run-llama/llama_index}} as the retrieval method and utilize PubMed and search results from Google as data sources to fetch them.

Based on CRAB, we evaluate the mainstream LLMs and explore potential improvements of biomedical curation based on Llama3 \citep{llama3}. \ignore{Considering the nature of open-ended queries, in which different LLMs may generate diverse responses using the same references, traditional QA metrics such as Accuracy and Exact Match, which are based on predefined answer texts, are not applicable.} In order to evaluate the curation ability more flexibly, we propose a citation-based evaluation method.
We frame the curation evaluation into a citation-based verification from two aspects: whether the retrieval-augmented LLMs can cite relevant references and whether it is unaffected by irrelevant references, directly quantifying the curation of retrieval-augmented LLMs in biomedicine.
In addition, we evaluate the latest reasoning LLMs and analyze the impact of explicit Chain-of-Thought on the biomedical curation of retrieval-augmented LLMs.
\ignore{Moreover, we conduct human evaluation to validate the rationale of this evaluation method.} 

Generally, our contributions are three-fold:
\begin{itemize}
    \item We introduce \textbf{CRAB} \footnote{Source data and code will be released after the paper is accepted.}, the first multilingual benchmark designed for the curation evaluation of retrieval-augmented LLMs in the domain of biomedicine.
    \item We formulate the evaluation of LLMs into a citation-based verification, quantitatively assessing the curation of retrieval-augmented LLMs, and perform human evaluations to substantiate the reliability.
    \item We conduct the evaluations of the mainstream LLMs on \textbf{CRAB} and validate the improvements of the biomedical curation based on Llama 3, highlighting the utility of \textbf{CRAB} for future research.
\end{itemize}

\section{CRAB}

In this section, we first introduce the curation ability of retrieval-augmented LLMs in (\ref{3.1}), which is the core ability we evaluate. Next, we outline our benchmark construction procedure in (\ref{3.2}) and peresent the proposed evaluation method in (\ref{3.3}).

\subsection{Curation}
\label{3.1}
Curation refers to the ability of retrieval-augmented LLMs to identify and cite relevant references when augmenting references, and to ignore irrelevant ones. In retrieval-augmented scenarios, LLMs combine internal knowledge with external retrieval knowledge, grounding their generated responses in external references. The curation process involves identifying, selecting, and citing pertinent references from a vast corpus of retrieved references based on their relevance and reliability\ignore{, thereby ensuring the precision and scalability of generated answers}.

In the domain of biomedicine, the importance of curation is particularly pronounced due to the critical nature of accurate and evidence-based information. Biomedical research and clinical decision-making rely heavily on trustworthy, validated findings. Consequently, the ability of retrieval-augmented LLMs to carefully curate references directly impacts their utility in clinical support systems, development of biomedical products, and patient information dissemination. Effective curation mitigates misinformation, enhances the interpretability of generated content, and facilitates transparent knowledge attribution, thereby strengthening the trustworthiness and practical applicability of LLMs in the biomedical domain.

\subsection{Data construction}
\label{3.2}

\begin{figure}[htbp]
    \centering
    \includegraphics[width=\linewidth]{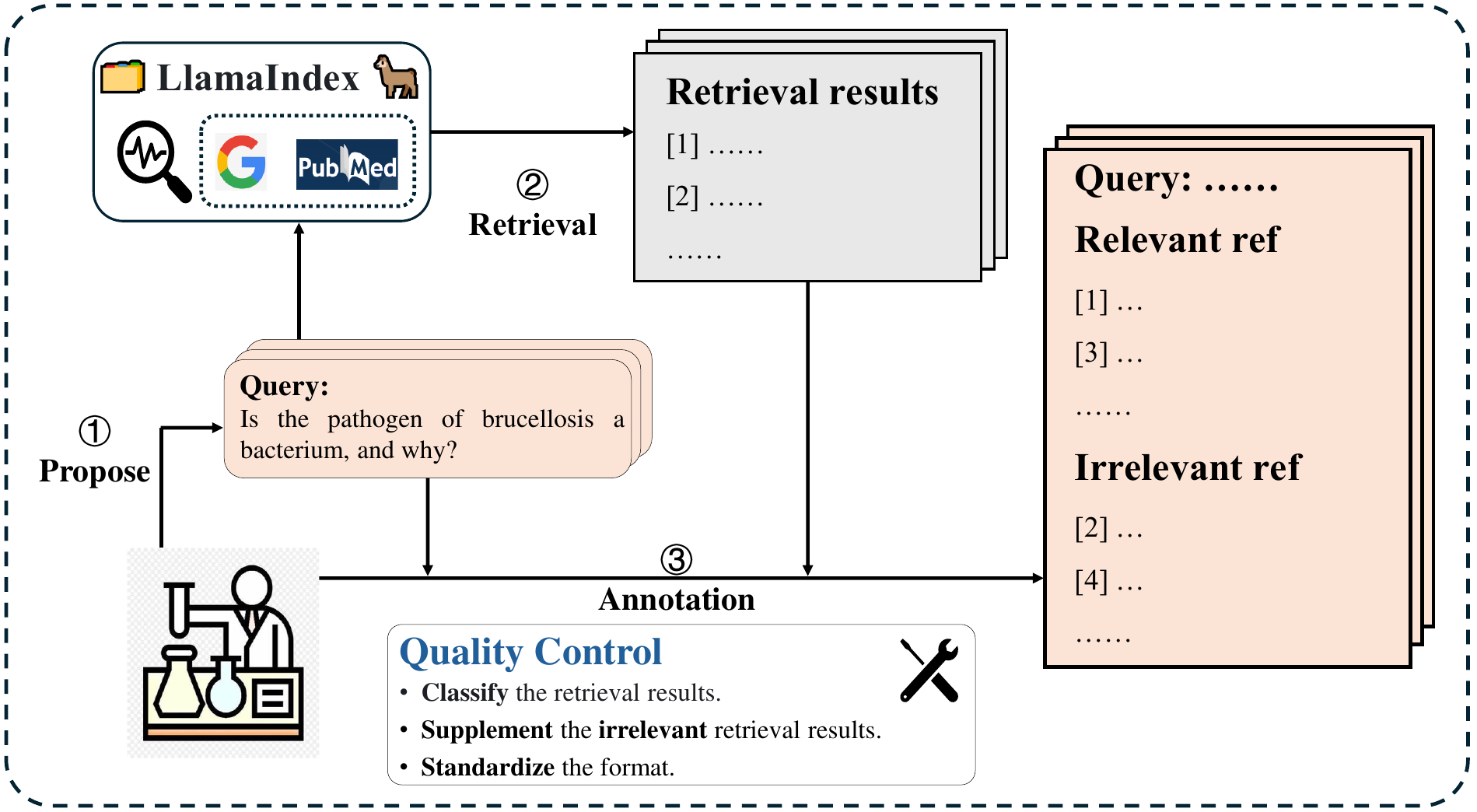}
    \caption{Data construction procedures of CRAB.}
    \label{data construction}
\end{figure}

\paragraph{Framework of CRAB} The overview of data construction for CRAB is shown in Figure \ref{data construction}. 
Unlike previously biomedical reference-based benchmarks \citep{mirage,comprehensivepracticalevaluationretrievalaugmented}, which typically provide pre-defined correct answer texts, CRAB focuses on open-ended biomedical queries without predefined ground-truth responses. In our framework, the correctness and appropriateness of LLM-generated answers depend entirely on the provided references. To rigorously evaluate the curation of retrieval-augmented LLMs, CRAB includes an adjustable setting, allowing controlled variation in the number of relevant and irrelevant references presented to the LLMs. Evaluation is conducted by verifying the relevance and correctness of the references cited within the generated answers, thereby providing a direct and robust measure of retrieval-augmented LLMs' proficiency in biomedical curation.

\begin{figure}[htbp]
    \centering
    \includegraphics[scale=0.3]{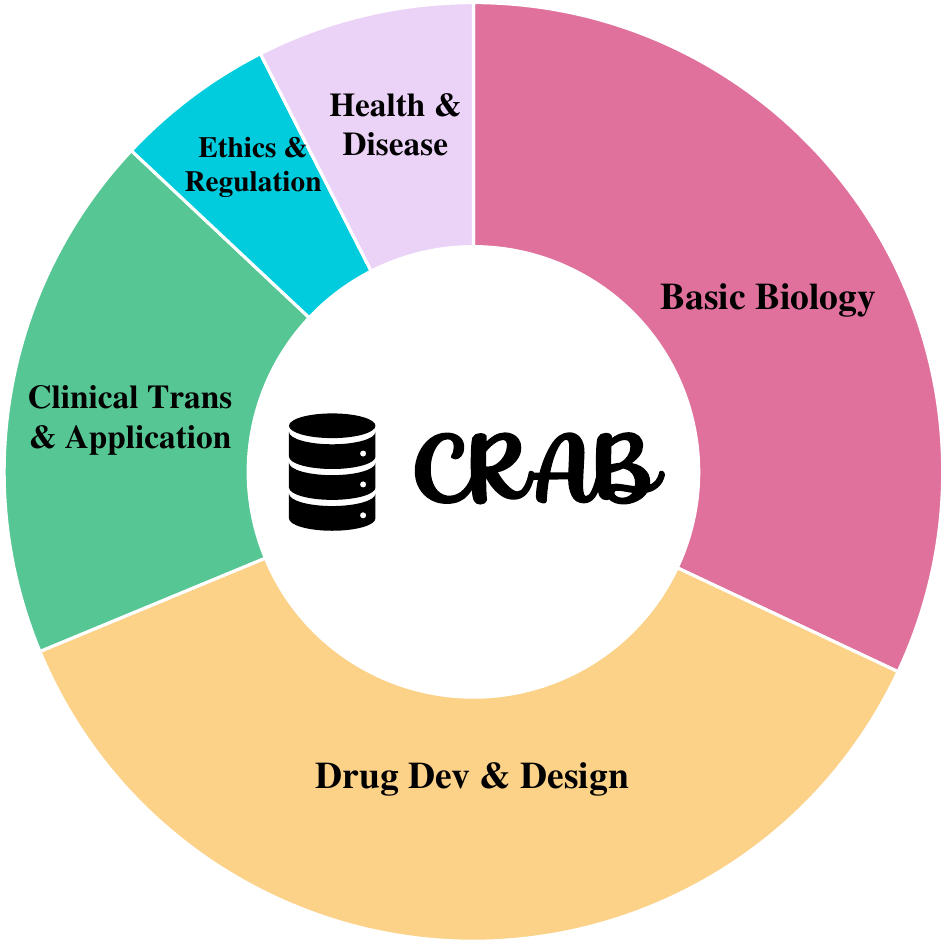}
    \caption{Distribution of query categories in CRAB}
    \label{distribution}
\end{figure}

\paragraph{Query collection} Instead of prompting LLMs to generate questions from open-source datasets \cite{mirage,comprehensivepracticalevaluationretrievalaugmented} and based on web data \citep{RGB}, we collect open-ended biomedical queries from experts \footnote{PhD graduates in biomedical major, and are paid \$1 for every five pieces of annotation.}. In detail, we collecte five categories of biomedical queries to reflect the performance in different dimensions: \textbf{Basic Biology}, \textbf{Drug Development and Design}, \textbf{Clinical Translation and Application}, \textbf{Ethics and Regulation}, \textbf{Public Health and Infectious Disease}. The distribution of query categories is shown in the Figure \ref{distribution}. 

\paragraph{Reference collection} After obtaining the biomedical queries, we apply LlamaIndex as the retireval method to get relevant references from PubMed and search results from Google. Then experts are asked to categorize the retrieved results into two sets: 1) content that can answer the query, i.e., relevant; 2) content that cannot answer the query, i.e., irrelevant. 
It is worth noting that for some queries, there are relatively few retrieval results that are irrelevant, or the degree of irrelevance is quite high. In order to reflect the real scenarios, we have supplemented them with high-quality irrelevant references. Specifically, these queries are reconstructed by replacing some biomedical entities within them. For example, the query ``The mechanism of action of Oseltamivir'' is reconstructed into "The indications of Oseltamivir". Then, the reconstructed queries are searched, and from the retrieval results, references are obtained that could not answer the original queries but have a certain degree of relevance to them. Since there may be references in the retrieval results of the reconstructed query that can answer the original query, the acquisition process is completed by experts, who conduct the classification of the second round of references. Therefore, we supplement some queries with high-quality irrelevant references to better evaluate the curation.
\ignore{For queries with a small number of irrelevant content, we perform \textbf{query rephrasing} (e.g., rewriting \enquote{the action mechanism of drug A} into \enquote{the adverse reactions of drug A}), and then use the new query to get references. The references of the rephrased queries are used as \textbf{noisy references}, and annotators are asked to conduct \textbf{a second round of classification} since the references of the rephrased queries might still be able to answer the original query. The reason of taking this approach is that in real-world applications, as retrieval systems continue to improve, the likelihood of completely irrelevant content appearing in the retrieved results is minimal and noise robustness should primarily target highly confusing noise content such as different attributes of the same entity.}

In CRAB, the format of drug, patent, paper, and clinical references is standardized, like the example shown in Figure \ref{noise example}, completely different from the document snippets extraction approaches \citep{RGB,mirage,comprehensivepracticalevaluationretrievalaugmented} and is more suitable for the evaluation of biomedical curation. Additionally, content-based deduplication on the references is performed to avoid the evaluation being too easy. At this point, we obtain both relevant and irrelevant sets of augmented references for each query. 
\ignore{Also, the language of queries and retrieved results in our benchmark maybe different because of the regional limitations in some researches. This simultaneously raises the language proficiency requirements for LLMs. }

\begin{table}[htbp]
    \centering
    \begin{tabular}{l|ccc}
    \hline
        \textbf{Language} & \textbf{query} & \textbf{\# pos.} & \textbf{\# neg.} \\
    \hline
        English & 100 & 622 & 462 \\
        French & 100 & 620 & 457 \\
        German & 100 & 615 & 450 \\
        Chinese & 100 & 610 & 485 \\
    \hline
    \end{tabular}
    \caption{The basic statistics of CRAB. Here \textbf{\# pos.} and \textbf{\# neg.} stand for relevant references and irrelevant references, respectively.}
    \label{statistics}
\end{table}

In terms of quantity, we collect \textbf{400} biomedicine queries and \textbf{2,467} relevant references and \textbf{1,854} irrelevant references as shown in the Table \ref{statistics}. In detail, We retain several queries without relevant references to simulate scenarios where relevant content might not be retrievable in real-world situations. For the other queries, we ensure each one has \textbf{at least two} relevant references and \textbf{three} irrelevant references to facilitate dynamic adjustments in the evaluation settings. Moreover, rather than relying solely on traditional query-level evaluation, CRAB shifts the evaluative focus to the \textbf{reference level}, significantly reducing the number of queries needed in evaluation. Specifically, the \textbf{400} evaluations on response content are transformed into more than \textbf{4,000} evaluations on reference citations.

\begin{figure}[h]
    \centering
    \includegraphics[width=\linewidth]{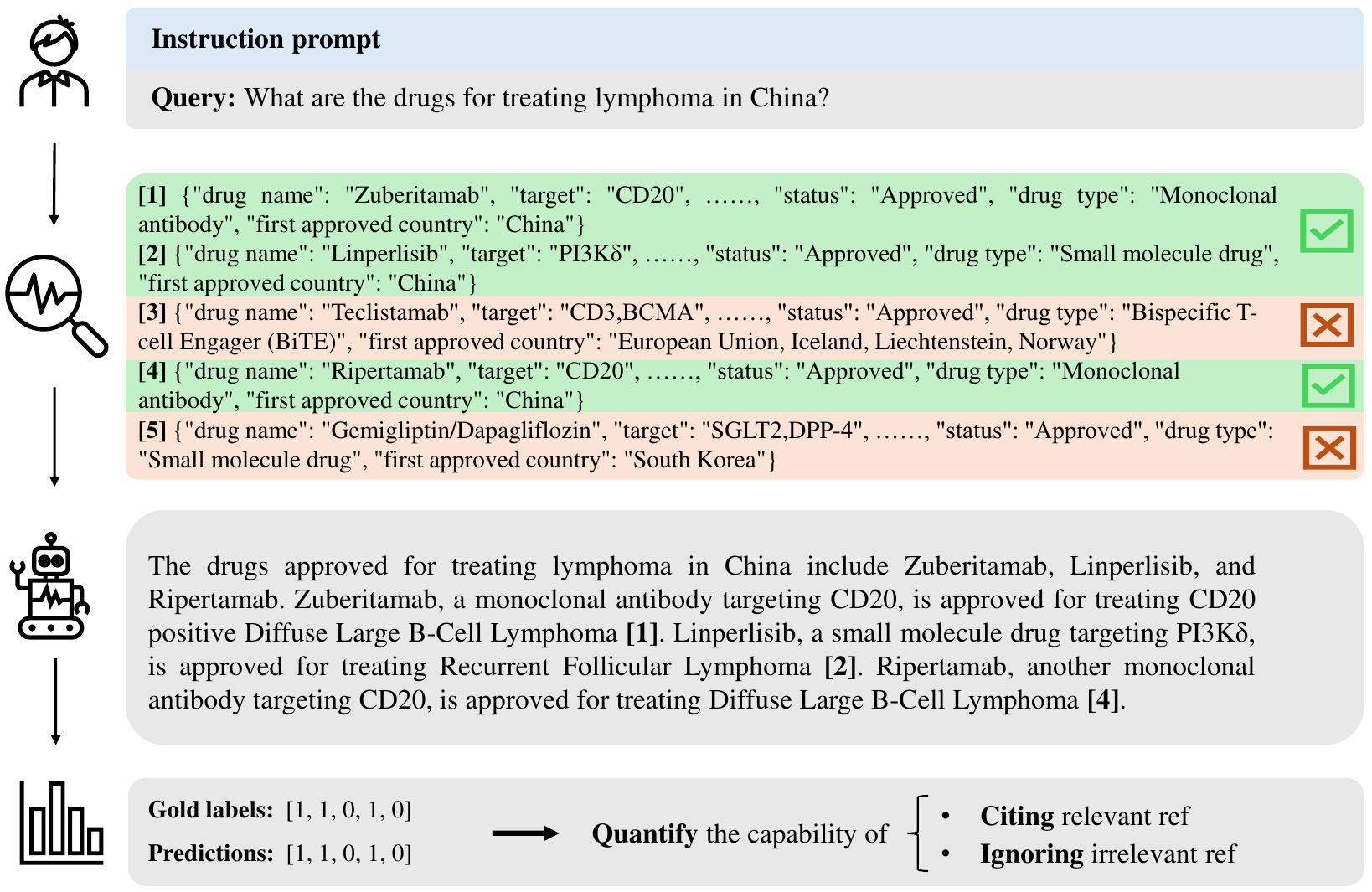}
    \caption{Our proposed evaluation method. Here green and check marks indicate relevant, while red and cross marks indicate irrelevant. The citations in the responses are regarded as the comprehensive processing results of the model on references.}
    \label{evaluation}
\end{figure}

\begin{figure*}[t]
    \centering
    \includegraphics[width=\linewidth]{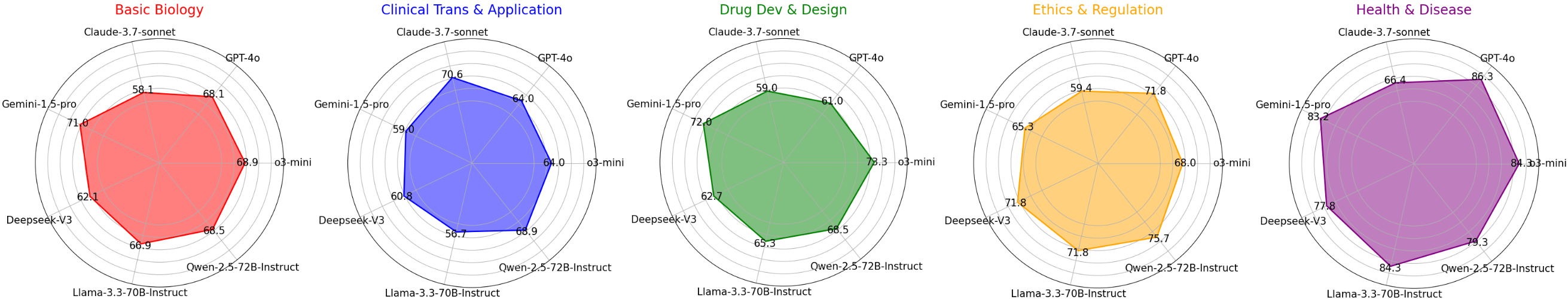}
    \caption{Performance comparison of LLMs across different query categories in English.}
    \label{class analysis}
\end{figure*}

\subsection{Evaluation method}
\label{3.3}
The core objective of CRAB is to assess the curation ability of retrieval-augmented LLMs. Specifically, CRAB examines whether retrieval-augmented LLMs successfully integrate relevant references while omitting irrelevant ones.
\ignore{The core purpose of CRAB is to evaluate the curation whether LLMs can correctly understand the relationship between queries and augmented references, utilize the relevant and remain unaffected by the irrelevant in their responses. We convert the original query-level response evaluation into a evaluation of \textbf{query-reference pairs classification}. }

Specifically, aiming at the demand for traceability of response in the biomedical domain, we model the evaluation of curation ability as a \textbf{citation-based verification}. It is based on three dimensions: the accuracy in identifying and citing relevant references (Relevance Precision, \textbf{RP}), the effectiveness in ignoring irrelevant references (Irrelevance Suppression, \textbf{IS}), and the overall curation performance (Curation Efficiency, \textbf{CE}). In terms of implementation, we characterize RP, IS, and CE by the F1 classification performance of two categories (relevant and irrelevant) and the Macro-avg F1 performance, respectively. For example, in Figure \ref{evaluation}, the 1st, 2nd, and 4th references are relevant, while the 3rd and 5th are irrelevant references. Therefore, in this case, the ground-truth labels are [1, 1, 0, 1, 0]. And the retrieval-augmented LLM cites the 1st, 2nd, and 4th references in the response, so the predicted labels are [1, 1, 0, 1, 0]. 
In this way, the evaluation of biomedical curation for retrieval-augmented LLMs focuses on the verification of the comprehensive processing results of references, rather than on the response content which neglects the evaluation of curation ability.

In addition, it must be mentioned that our proposed evaluation method does not serve as a replacement for content-based evaluations. Instead, it constitutes an independent evaluation dimension that operates alongside traditional content evaluations. Moreover, this citation-based approach can function as a preliminary assessment, providing an a priori measure that informs and complements subsequent in-depth analyses of response content.

To verify the reliability of our evaluation method, we conduct the human evaluation of the results in the English section. For each citation in the response, the evaluators (biomedical experts in the annotating phase) are required to check whether the content preceding it is generated based on the corresponding reference.

\section{Experiments}
We conduct an evaluation of curation on CRAB for existing mainstream general LLMs such as GPT-4o, Gemini-1.5-pro, etc., and mainstream reasoning LLMs such as o3-mini, DeepSeek-R1, etc., and analyze the evaluation results from the five categories of data in CRAB. Moreover, we verify the potential for improvement of curation based on Llama3 and validate the reliability of the proposed evaluation metric through human evaluation.

\subsection{Experimental settings}
\textbf{Task formats.} Given the high cost of large-parameter evaluation, for each query, we randomly select \textbf{2 relevant} references and \textbf{3 irrelevant} references from the annotated sets of references. We fix the random seed at 42 to sample augmented references and \textbf{shuffle} the order of the references before generation \citep{lost-in-the-middle}. For each model, we use the official recommended generation hyperparameters.

\noindent \textbf{Models.} The LLMs used in the experiment include closed-source models such as GPT and open-source models such as Llama.

\subsection{Experimental results}
\label{experimental results}
We evaluate the mainstream LLMs on CRAB with the citation-based evaluation metric introduced in \ref{3.3} and the main performance can be seen in Table \ref{main performance} (the overall performance can be seen in Appendix \ref{overall performance}). We report the performance from three aspects: the accuracy in identifying and referencing relevant sources (Relevance Precision, \textbf{RP}), the effectiveness in ignoring irrelevant sources (Irrelevance Suppression, \textbf{IS}), and the overall curation performance (Curation Efficiency, \textbf{CE}). The performance of \textbf{RP} and \textbf{IS} reflects the ability to identify and integrate relevant and irrelevant references in curation, respectively. \textbf{CE}, on the other hand, reflects the overall curation ability.

\begin{table*}
    \centering
    \begin{adjustbox}{max width=\textwidth}
    \begin{tabular}{l|ccc|ccc}
    \hline
    \multirow{2}{*}{\textbf{Models}} & \multicolumn{3}{c|}{\textbf{English}} & \multicolumn{3}{c}{\textbf{French}} \\
     &  \textbf{RP} (F1, \%) & \textbf{IS} (F1, \%) & \textbf{CE} (F1, \%) &  \textbf{RP} (F1, \%) & \textbf{IS} (F1, \%) & \textbf{CE} (F1, \%) \\
     \hline
     GPT-4-turbo & 67.10 & 71.46 & 69.28 & 58.58 & 78.46 & 68.52 \\
     GPT-4o & 67.13 & 74.14 & 70.63 & 65.50 & 76.53 & 71.02 \\
     Claude-3.7-sonnet & 64.19 & 57.02 & 60.60 & 69.69 & 70.78 & 70.23 \\
     Gemini-1.5-pro & 64.86 & \textbf{78.96} & \textbf{71.91} & 57.14 & 76.49 & 66.82 \\
     DeepSeek-V3 & 61.26 & 64.82 & 64.82 & 62.86 & 76.29 & 69.57 \\
     Doubao & \textbf{74.02} & 61.65 & 67.84 & 61.45 & \textbf{79.32} & 70.38 \\
     DeepSeek-R1 & 61.65 & 50.23 & 55.94 & 66.27 & 65.30 & 65.78 \\
     QwQ-32B & 63.02 & 57.21 & 60.11 & 68.00 & 73.23 & 70.62 \\
     Gemini-2.0-thinking & 67.26 & 73.06 & 70.16 & \textbf{69.32} & 76.65 & \textbf{72.98} \\
     o3-mini & 69.16 & 73.78 & 71.47 & 68.60 & 77.35 & \textbf{72.98} \\
     \hline
     \multicolumn{7}{c}{} \\[-10pt]
    \hline
     \multirow{2}{*}{\textbf{Models}} & \multicolumn{3}{c|}{\textbf{German}} & \multicolumn{3}{c}{\textbf{Chinese}} \\
     &  \textbf{RP} (F1, \%) & \textbf{IS} (F1, \%) & \textbf{CE} (F1, \%) &  \textbf{RP} (F1, \%) & \textbf{IS} (F1, \%) & \textbf{CE} (F1, \%) \\
     \hline
     GPT-4-turbo & 56.29 & 77.68 & 66.98 & 70.40 & 82.01 & 76.20 \\
     GPT-4o & 63.73 & 76.74 & 70.24 & 76.04 & \textbf{84.87} & \textbf{80.46} \\
     Claude-3.7-sonnet & 69.69 & 70.78 & 70.23 & 71.57 & 72.37 & 71.97 \\
     Gemini-1.5-pro & 57.14 & 76.49 & 66.82 & 62.57 & 78.86 & 70.72 \\
     DeepSeek-V3 & 61.98 & 75.83 & 68.90 & 71.28 & 82.47 & 76.87 \\
     Doubao & 60.18 & \textbf{79.20} & 69.69 & \textbf{82.08} & 67.98 & 75.03 \\
     DeepSeek-R1 & 67.64 & 69.55 & 68.59 & 68.97 & 69.34 & 69.15 \\
     QwQ-32B & 65.25 & 68.22 & 66.74 & 64.99 & 61.89 & 63.44 \\
     Gemini-2.0-thinking & 68.84 & 75.99 & 72.41 & 72.85 & 79.14 & 76.00 \\
     o3-mini & \textbf{70.28} & 77.66 & \textbf{73.97} & 73.16 & 80.21 & 76.68 \\
    \hline
    \end{tabular}
    \end{adjustbox}
    \caption{The main performance of representative LLMs}
    \label{main performance}
\end{table*}

Evaluation results are shown in Table \ref{main performance}. The experimental results demonstrate distinct language-specific performance patterns. Notably, Gemini-1.5-pro exhibits superior performance in English, achieving a \textbf{CE} F1 score of 71.91. In contrast, o3-mini demonstrates peak performance in Romance languages, attaining \textbf{CE} F1 scores of 72.98 for French and 73.97 for German. Meanwhile, GPT-4o records its optimum results in Chinese, reaching an exceptional \textbf{CE} F1 score of 80.46. From the perspective of language comparison, it can be observed from the Table \ref{main performance} that the latest closed-source LLMs perform more consistently compared to open-source LLMs, primarily due to their larger parameter sizes and more extensive multilingual pretraining knowledge. Notably, each LLM exhibits biases towards either \textbf{RP} or \textbf{IS}. For example, Doubao achieves the highest \textbf{RP} performance in the English section, but its \textbf{IS} performance ranks among the last few among the closed-source LLMs. The evaluation results show that there are significant differences in the curation ability of each LLM to identify and integrate references in the biomedical domain. In this way, we quantify the curation of retrieval-augmented LLMs in biomedical domain.

We also evaluate the recent popular reasoning LLMs, as shown in the last four lines of each section in Table \ref{main performance}. We select four representative reasoning LLMs for evaluation and analysis, namely o3-mini, Gemini-2.0-thinking, DeepSeek-R1, and QwQ. From the evaluation results, it can be seen that the DeepSeek-R1 and QwQ have not improved in terms of biomedical curation compared to their original base models (QwQ is trained based on Qwen-2.5-32B). Instead, in both the English and Chinese sections, the performance of both has declined. 
We attribute this to the excessive training of the reasoning ability of the two LLMs in mathematical and code, which leads to overthinking in the domain of biomedicine and blurs the boundaries of professional concepts in biomedicine. We provide detailed examples to analyze this phenomenon in \ref{Case study of reasoning LLMs}. 
On the contrary, o3-mini and Gemini-2.0-thinking basically maintain their leading positions compared with their homologous LLMs.

To provide a more comprehensive comparison, we conduct evaluations of the seven representative large-parameter LLMs across each query category in the English section, as \textbf{CE} F1 scores shown in the Figure \ref{class analysis}. As can be seen, the best-performing LLM varies across different categories. This also reflects the complexity of the biomedical domain compared with the general domain, and proves that the biomedical domain requires the evaluation of the curation ability of retrieval-augmented LLMs.

\subsection{Verification of improvements}
To verify the potential for improvement of biomedical curation, we conduct a validation based on Llama3-70B, details can be seen in Appendix \ref{verification}.

\subsection{Human evaluation}
In order to verify the reliability of the evaluation method based on citation results proposed by us, we conducted a manual evaluation, specifically evaluating the results of Gemini-1.5-pro, which performed the best among non-reasoning-capable LLMs in the English section. Evaluators (biomedical experts in the annotating phase) are asked to verify whether there is corresponding reference content for the text before the citation number in the response, so as to exclude the situation where the citation is inconsistent with the content. 

\begin{table}[htbp]
    \centering
    \begin{tabular}{l|ccc}
       \hline
        \textbf{Settings} & \textbf{RP} & \textbf{IS} & \textbf{CE} \\
        \hline
        \textbf{citation-based} & 64.86 & 78.96 & 71.91 \\
        \textbf{human} & 64.85 & 79.23 & 72.04 \\
        \hline
    \end{tabular}
    \caption{Comparison between citation-based and human evaluation}
    \label{human evaluation results}
\end{table}

The comparison between the citation-based evaluation and the human evaluation is shown in Table \ref{human evaluation results}. It can be seen that the F1 scores of \textbf{RP}, \textbf{IS} and \textbf{CE} are basically consistent between them. Evaluators find that model sometimes explains the reasons for not citing irrelevant references, and they judge this as correct identification of relevance. Therefore, the performance of the human evaluation is relatively higher. Overall, there is almost no gap between the two, verifying the effectiveness of CRAB and our proposed evaluation method.

\section{Analysis}
We conduct a detailed analysis of the evaluation results. Firstly, We analyze the reasons for the performance degradation of some reasoning LLMs. Secondly, we conduct an analysis of overlap biomedical entities in the query, reference, and response. Lastly, we analyze the type and quantity of errors. 

\section{Conclusion}
In this paper, we introduce the the benchmark for \textbf{C}uration of the \textbf{R}etrieval-\textbf{A}ugmented LLMs in \textbf{B}iomedicine (\textbf{CRAB}) and evaluate the curation of the retrieval-augmented LLMs in the biomedical domain. To conduct the evaluation, we propose a citation-based evaluation method to quantify the curation. In addition, we verify the potential of improvements on curation. Evaluation results demonstrates the obvious gaps in biomedical curation among different retrieval-augmented LLMs. 

\section{Limitations}
Our work focuses on evaluating the curation of retrieved-augmented LLMs, and we have not yet proposed systematic methods for improving curation capabilities. We will focus on enhancing curation capabilities in our future work.



\bibliography{custom}

\clearpage
\appendix

\section*{Appendix}
\section{Related Work}

Large Language Models (LLMs) excel in text generation but struggle with outdated knowledge, domain-specific gaps, and hallucinations \citep{hallucination2023,factuality,hallucination2024}. Retrieval-Augmented Generation (RAG) \citep{RAG} mitigates these issues by retrieving external knowledge to improve responses \citep{yu2024RAG,gao2024RAG,huang2024RAG}. Currently, the evaluation of retrieval-augmented LLMs focuses on the general domain, verifying whether they can correctly answer questions with augmented references \cite{RAAT,RGB,nomiracl,explainable}.
In the biomedical domain, the evaluation of retrieval-augmented LLMs assumes heightened importance relative to the general domain due to the critical nature of clinical decision-making and research integrity.

\paragraph{Biomedical Benchmark} Initially, benchmarks without the augmented references were the focus\citep{jin2021disease,hendrycks2020measuring,jin2019pubmedqa,liu2023benchmarking,dahl,MedExpQA}. For example, \citet{dahl} introduced DAHL, a benchmark for assessing hallucinations in long-form text generation. It contains thousands of questions from biomedical research papers, evaluating fact-conflicting hallucinations by deconstructing responses into atomic units and calculating the DAHL Score. \citet{MedExpQA} presented MedExpQA, the first multilingual benchmark for medical Question Answering. It includes gold reference explanations written by medical doctors, allowing for a more comprehensive evaluation of LLMs' reasoning abilities across different languages. However, due to issues like hallucinations and outdated knowledge in LLMs, recent work in the biomedical domain has increasingly focused on developing benchmarks to evaluate retrieval-augmented LLMs, recognizing that the integration of relevant references is critical for enhancing the transparency and reliability. 
\citet{mirage} proposed MIRAGE, augmenting references with multi-choice questions from five medical QA datasets to systematically evaluate the performance of retrieval-augmented LLMs. 
In addition, \citet{comprehensivepracticalevaluationretrievalaugmented} proposed MedRGB by using questions from four medical QA datasets, generating retrieval topics, conducting offline and online retrieval, and constructing four test scenarios to comprehensively evaluate the performance of medical retrieval-augmented LLMs. 
However, previous benchmarks neglected the evaluation of biomedical curation, which enhances transparency and reliability of the retrieval-augmented LLMs. Therefore, we propose CRAB to evaluate the bimedical curation of retrieval-augmented LLMs.

\paragraph{Metrics} Most metrics for the evaluation of retrieval-augmented LLMs focus on determining whether the response contains the pre-defined answer \citep{RGB,mirage}. Fine-tuned retrieval-augmented LLMs also use metrics like exact match and text utilization to verify improvements brought by training and augmented references \citep{RAAT,explainable}. Moreover, \citep{enabling_citation} proposes a citation-based metric relying on an NLI model to score fluency, correctness, and citation quality of short QA tasks. RAGChecker \citep{ragchecker} breaks down responses into correct and incorrect claims, assessing context utilization, noise sensitivity, hallucination, and faithfulness.
RAGAS \citep{ragas} and ARES \citep{ares} adopt the RAG Triad framework \citep{rag-triad}, evaluating context relevance, groundedness, and answer relevance.
However, previous metrics focused on predefined rules or the approach of "LLM-as-judge," and they are not applicable to the evaluation of curation. To address this, we propose a citation-based verification metric to quantify the biomedical curation of retrieval-augmented LLMs.

\section{Analysis}
\label{analysis}
We conduct a detailed analysis of the evaluation results. Specifically, we first analyze the reasons for the performance degradation of the reasoning LLMs mentioned in \ref{experimental results}. Secondly, we carry out an overlap analysis of biomedical entities in the query, reference, and response. Finally, we conduct an error analysis and count the types of errors. 

\begin{figure*}[htbp]
    \centering
    \includegraphics[width=\linewidth]{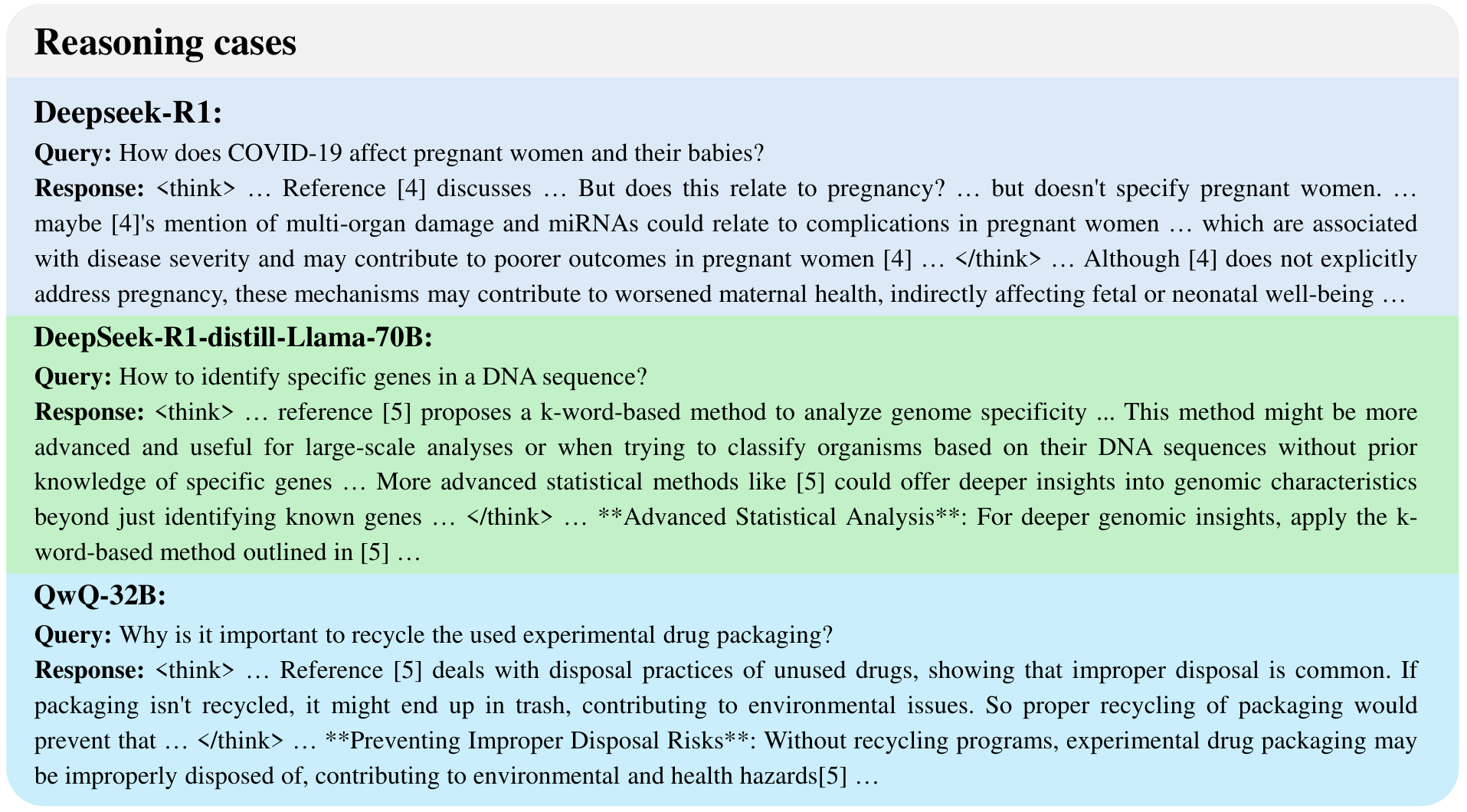}
    \caption{Cases of reasoning LLMs}
    \label{Cases of reasoning LLMs}
\end{figure*}

\subsection{Case study of reasoning LLMs}
\label{Case study of reasoning LLMs}
In order to understand the reasons why the biomedical curation of the DeepSeek-R1 series and the QwQ reasoning LLMs have declined compared to their homologous LLMs, we analyzed their incorrect cases, as shown in Figure \ref{Cases of reasoning LLMs}. 

Since the LLMs in Figure \ref{Cases of reasoning LLMs} obtain their reasoning capabilities through rule-based reinforcement learning based on a large amount of mathematical and code data, the situation of overthinking has occurred in the domain of biomedicine. For example, in the case of DeepSeek-R1, reference [4] discusses cardiac metabolic microRNAs and their relation to COVID-19 severity and mortality. However, due to its continuous questioning of its own judgment on the relevance, it eventually identified this reference as relevant to answering the query ``How does COVID-19 affect pregnant women and their babies?''. In addition, in the cases of DeepSeek-R1-distill-Llama-70B and QwQ-32B, ``Statistical specificity analysis method based on k-words'' and ``Methods for the storage and disposal of unused and expired drugs'' are excessively speculated in the process of reasoning, to be able to answer the queries ``How to identify specific genes in a DNA sequence?'' and ``Why is it important to recycle the used experimental drug packaging?''. In the responses, they are described as ``Advanced Statistical Analysis'' and ``Preventing Improper Disposal Risks''.  

\subsection{Entity-based analysis}
\label{entity-based analysis}
In order to have a deeper understanding of the evaluation results of biomedical curation, we conduct a quantitative analysis based on the biomedical entity set in the query, reference, and answer, named $E_q$, $E_r$, $E_a$. We analyze from four aspects: 1). The proportion of entities in $E_a$ that are covered in $E_r$, called \textbf{Entity Precision} (\textbf{EP}); 2). The proportion of entities in $E_r$ that are covered in $E_a$, called \textbf{Entity Recall} (\textbf{ER}); 3). Using the Jaccard similarity to measure the similarity between $E_a$ and $E_r$, called \textbf{Jaccard}; 4). The proportion of entities in $E_q$ among the entity list of the response, called \textbf{Query Entity Coverage} (\textbf{QEC}). Considering that most of the currently open-source biomedical NER models are in English, we use the open-source model BioMed\_NER \footnote{\url{https://huggingface.co/Helios9/BioMed_NER}} and biomedical-ner-all \footnote{\url{https://huggingface.co/d4data/biomedical-ner-all}} to analyze seven large-parameter LLMs in the English section.

\begin{figure*}[ht]
    \centering
    \includegraphics[width=\linewidth]{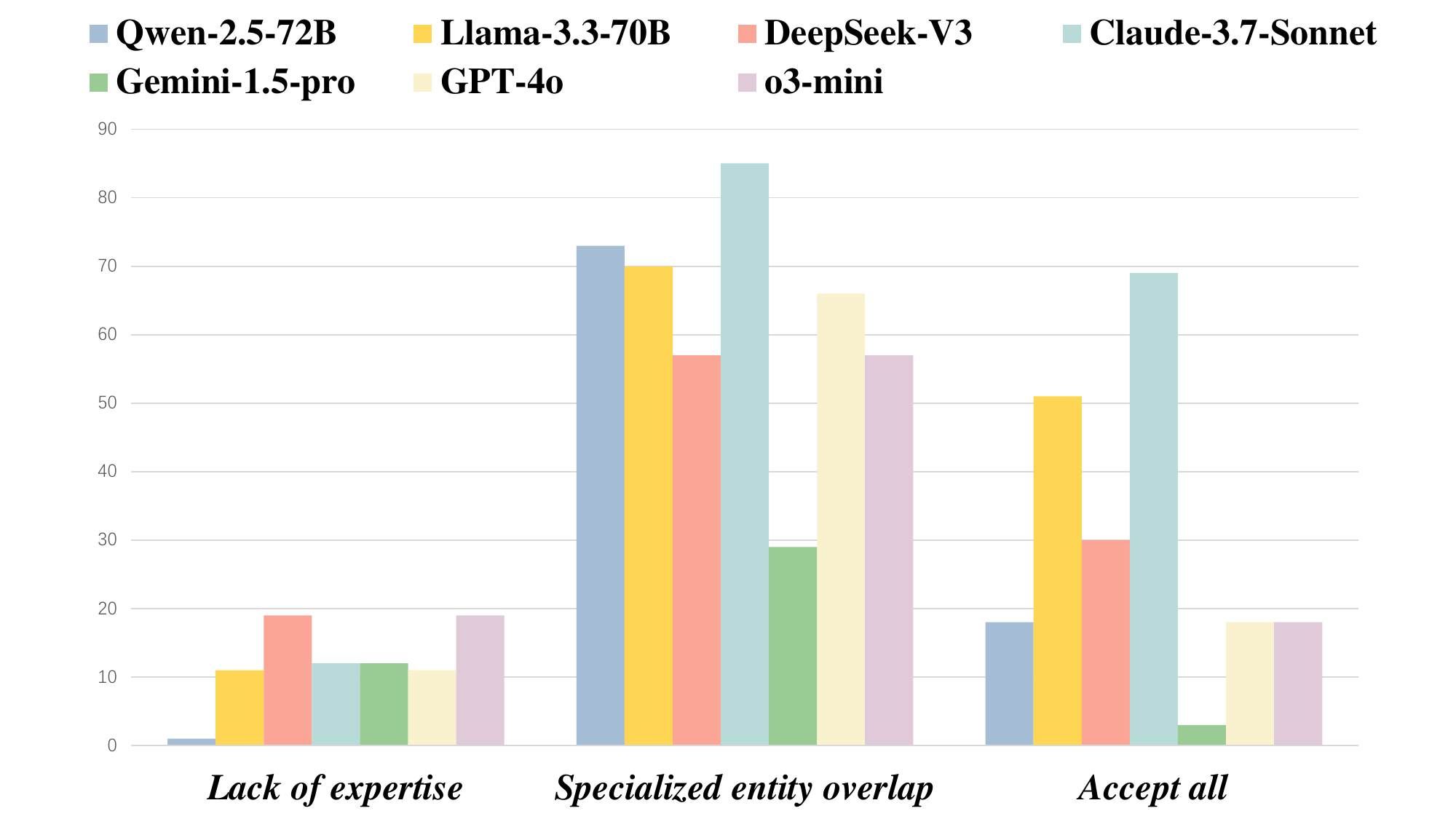}
    \caption{Statistics on the number of citation error types.}
    \label{error_types}
\end{figure*}

\begin{table}[htbp]
    \centering
    \begin{tabular}{l|ccccc}
        \hline
        \textbf{Models} & \textbf{EP} & \textbf{ER} & \textbf{Jaccard} & \textbf{QEC} \\
        \hline
        GPT-4o & 23.68 & 4.97 & 4.28 & 7.19 \\
        Gemini-1.5 & 32.59 & 7.51 & 6.50 & 11.28 \\
        Claude-3.7 & 31.15 & 8.21 & 6.95 & 8.06 \\
        DeepSeek & 25.81 & 5.53 & 4.77 & 10.30 \\
        Llama-3.3 & 30.70 & 7.51 & 6.42 & 16.53 \\
        Qwen-2.5 & 28.60 & 6.04 & 5.25 & 9.26 \\
        o3-mini & 20.86 & 6.09 & 4.95 & 8.32 \\
        \hline
    \end{tabular}
    \caption{Results of entity-based analysis}
    \label{results of entity-based analysis}
\end{table}

\textbf{EP}, \textbf{ER}, and \textbf{Jaccard} reflect whether the answering tendencies of different retrieval-augmented LLMs are more based on references or more based on their internal knowledge. It is worth noting that there is no such thing as the higher or the lower the better for these three dimensions. Because if the values are too high, the LLM will mindlessly state the content of the reference, and if they are too low, the LLM will deviate from the content of the reference. Their value lies in the comparative analysis of the reasons for the curation differences among different LLMs, rather than serving as independent metrics. For example, Claude-3.7-sonnet has a relatively lower performance in the curation evaluation compared to the other six LLMs, but it is significantly higher than the other LLMs in terms of \textbf{EP}, \textbf{ER}, and \textbf{Jaccard}. This reflects its lack of the ability to judge the relevance of references, and it has a greater tendency to directly use the explanations of references as the content of its answers.  

On the other hand, \textbf{QEC} reflects the faithfulness to the query. A lower \textbf{QEC} indicates that the LLM is more interfered by the irrelevant content in the references. This is also reflected in Claude-3.7-sonnet, whose \textbf{QEC} is lower.

\subsection{Error analysis}
\label{error analysis}

\begin{table*}[t]
    \centering
    \begin{tabular}{>{\centering\arraybackslash}m{3.5cm}|>{\centering\arraybackslash}m{5.5cm}|>{\centering\arraybackslash}m{5.5cm}}
        \hline
        \centering{\textbf{query}} & \centering{\textbf{\# neg.}} & \textbf{response}  \\
        \hline
        \multicolumn{1}{m{3.5cm}|}{What drugs are available for CD47 and PDL1 \textcolor{red}{\textbf{bispecific antibodies?}}} & \multicolumn{1}{m{5.5cm}|}{[4] \{\enquote{drug name}: \enquote{JY-207b}, \enquote{target}: \enquote{CD47,PDL1}, ...\ignore{\enquote{mechanism action}: \enquote{Cluster of differentiation 47 inhibitors,Programmed death-ligand 1 inhibitors}, \enquote{indications}: \enquote{Neoplasms}, }, \enquote{status}: \enquote{Preclinical}, \textcolor{red}{\textbf{\enquote{drug type}: \enquote{Antibody drug conjugate}}}\}} & \multicolumn{1}{m{5.5cm}}{Drugs that are dual antibodies against CD47 and PDL1 include IMM-2520, \textcolor{red}{\textbf{JY-207b}}, and BAT-7104 [1]\textcolor{red}{\textbf{[4]}}[5].} \\
        \hline
        \multicolumn{1}{m{3.5cm}|}{Who is more likely to get an Salmonella infection \textcolor{red}{\textbf{and severe illness}}?} & \multicolumn{1}{m{5.5cm}|}{[4] Get the Facts about Salmonella | FDA  People \textcolor{red}{\textbf{can also become infected}} with Salmonella by handling contaminated food\ignore{, and then accidentally spreading the bacteria from their hands to} ...} & \multicolumn{1}{m{5.5cm}}{\textcolor{red}{\textbf{Individuals handling contaminated food are more likely to get a Salmonella infection as the bacteria can be spread from their hands to their mouths [4]}}\ldots} \\
        \hline
    \end{tabular}
    \caption{Error cases where \textbf{\# neg.} stand for irrelevant references.}
    \label{error case}
\end{table*}

To better comprehend the curation of retrieval-augmented LLMs in the biomedical domain, we analyze the incorrect cases in the English section and classify errors into \textbf{three} categories: 

\textbf{(1) Lack of expertise.} Sometimes LLMs lack sensitivity to biomedicine knowledge, leading to the confusion of important concepts in biomedicine. For example, in the first case in Table \ref{error case}, LLM fails to differentiate between the bispecific antibodies and antibody-drug conjugates. \ignore{Therefore, it incorrectly identifies the drug \enquote{JY-207b} as a dual-specificity antibody drug targeting both CD47 and PDL1 and cite the irrelevant ``reference [4]''.}

\textbf{(2) Specialized entity overlap.} For the irrelevant references that contain the key entities of the query, the LLMs may confuse and explain them in responses. Take the second case in Table \ref{error case} as an instance, the query expresses which groups of people are more susceptible to Salmonella infection, while the reference explains that humans can contract it through contact with contaminated food. Both contain the key entity ``Salmonella'', and LLM forcibly uses the transmission mode to answer the susceptible populations.

\textbf{(3) Accept all.} LLMs somtimes generate responses to queries based solely on all references without considering their relevance. This situation mostly occurs in queries that tend to list information, such as \enquote{clinical studies of a certain drug}, where LLMs proceed to describe the provided references one by one.

Based on the analysis above, we manually conduct a count of these three types of errors for some LLMs in the English section, as shown in the Figure \ref{error_types}. To enhance the comprehension of LLMs, these three aspects are suitable starting points.

\section{Experimental Results}
\subsection{Human Evaluation}
\label{human evaluation}
In order to verify the reliability of the evaluation method based on citation results proposed by us, we conducted a manual evaluation, specifically evaluating the results of Gemini-1.5-pro, which performed the best among non-reasoning-capable LLMs in the English section. Evaluators (biomedical experts in the annotating phase) are asked to verify whether there is corresponding reference content for the text before the citation number in the response, so as to exclude the situation where the citation is inconsistent with the content. 

\begin{table}[htbp]
    \centering
    \begin{tabular}{l|ccc}
       \hline
        \textbf{Settings} & \textbf{pos.} & \textbf{neg.} & \textbf{macro avg} \\
        \hline
        \textbf{citation-based} & 64.86 & 78.96 & 71.91 \\
        \textbf{human} & 64.85 & 79.23 & 72.04 \\
        \hline
    \end{tabular}
    \caption{Comparison between citation-based and human evaluation}
    \label{human evaluation results}
\end{table}

The comparison between the citation-based evaluation and the human evaluation is shown in Table \ref{human evaluation results}. It can be seen that the F1 values are basically consistent between them. In the human evaluation, experts find that Gemini-1.5-pro sometimes explains the reasons for not citing irrelevant references, and judge this as a correct identification of relevance, setting the predicted label to 0. Therefore, the performance of the human evaluation is relatively higher than that of the citation-based evaluation.

\subsection{Verification of improvements}
\label{verification}
In addition, in order to verify the potential for improvement of the curation ability of retrieval-augmented LLMs in the domain of biomedicine, we conduct a validation experiment. In detail, we conduct Continual Pre-Training (CPT) of Llama3-70B with biomedical articles and perform Supervised Fine-Tuning (SFT) on approximate 2,000 synthesized biomedical QA data containing augmented references. Considering the high cost of CPT, we conduct validation experiments only in the Chinese section. Both CPT data and SFT data are collected from open-source biomedical data. The F1-scores of validation experiments are shown in Table \ref{validation}. 

\begin{table}[htbp]
    \centering
    \begin{tabular}{l|ccc}
    \hline
    \textbf{Models} & \textbf{RP} & \textbf{IS} & \textbf{CE} \\
    \hline
    Instruct & 69.21 & 77.49 & 73.35 \\
    + SFT & 64.93 & 74.04 & 69.48 \\
    + CPT\&SFT & 71.01 & 79.24 & 75.13 \\
    \hline
    \end{tabular}
    \caption{Results of F1 scores in validation experimentsbased on LLaMa3.}
    \label{validation}
\end{table}

The results of the verification experiment are shown in Table \ref{validation}. As can be seen from the results, after deepening the understanding of biomedical knowledge through CPT, the biomedical curation ability of Llama3 has been significantly improved compared with the version that only undergoes SFT and the official instruct version. It is worth noting that since the validation experiment just simply verifies the potential for improving biomedical curation by adding data related to biomedicine, and there is no comprehensive evaluation, so we do not propose a new biomedical LLM, nor can it draw conclusions such as ``CPT is necessary in the biomedical domain.''

\subsection{Overall Performance}
\label{overall performance}
We evaluate various parameter versions of mainstream LLMs on CRAB, and the evaluation results are shown in Table \ref{en french performance} and Table \ref{german cn performance}. 

\begin{table*}[htbp]
  \centering
  \begin{adjustbox}{max width=\textwidth}
\begin{tabular}{l|ccc|ccc|ccc}
    \hline
    \multirow{2}{*}{\textbf{Models}} & \multicolumn{3}{c|}{\textbf{RP}} & \multicolumn{3}{c|}{\textbf{IS}} & \multicolumn{3}{c}{\textbf{CE}} \\
     & P (\%) & R (\%) & F1 (\%) & P (\%) & R (\%) & F1 (\%) & P (\%) & R (\%) & F1 (\%) \\
    \hline
    \multicolumn{10}{c}{\textbf{English}} \\
    \hline
    GPT-3.5-turbo & 51.60 & 74.23 & 59.50 & 77.00 & 54.67 & 63.94 & 64.30 & 64.70 & 62.50 \\
    GPT-4-turbo & 58.11 & 79.38 & 67.10 & 82.53 & 63.00 & 71.46 & 70.32 & 71.19 & 69.28 \\
    GPT-4o & 60.58 & 75.26 & 67.13 & 81.03 & 68.33 & 74.14 & 70.80 & 71.80 & 70.63 \\
    GPT-4o-mini & 58.80 & 80.93 & 68.11 & 83.70 & 63.33 & 72.11 & 71.25 & 72.13 & 70.11 \\
    Claude-3.5-sonnet & 49.58 & 91.24 & 64.25 & 87.59 & 40.00 & 54.92 & 68.59 & 65.62 & 59.58 \\
    Claude-3.7-sonnet & 50.14 & 89.18 & 64.19 & 85.91 & 42.67 & 57.02 & 68.03 & 65.92 & 60.60 \\
    Gemini-1.5-pro & 68.18 & 61.86 & 64.86 & 76.73 & 81.33 & \textbf{78.96} & 72.46 & 71.59 & \textbf{71.91} \\
    DeepSeek-V3 & 54.40 & 70.10 & 61.26 & 76.23 & 66.05 & 64.82 & 65.31 & 66.05 & 64.82 \\
    Doubao-pro & 75.43 & 72.67 & \textbf{74.02} & 60.00 & 63.40 & 61.65 & 67.72 & 68.03 & 67.84 \\
    GLM-4-9B-chat & 55.00 & 51.03 & 52.94 & 69.75 & 73.00 & 71.34 & 62.37 & 62.02 & 62.14 \\
    Llama-3.1-8B-Instruct & 56.43 & 70.10 & 62.53 & 77.08 & 65.00 & 70.52 & 66.75 & 67.55 & 66.53 \\
    Llama-3.1-Tulu-3-8B & 73.56 & 21.33 & 33.07 & 42.01 & 88.14 & 56.91 & 57.79 & 54.74 & 44.99 \\
    Llama-3.1-70B-Instruct & 56.41 & 79.38 & 65.95 & 81.90 & 60.33 & 69.48 & 69.16 & 69.86 & 67.72 \\
    Llama-3.1-Nemotron-70B & 47.50 & 88.14 & 61.73 & 82.84 & 37.00 & 51.15 & 65.17 & 62.57 & 56.44 \\
    Llama-3.1-Tulu-3-70B & 47.12 & 92.78 & 62.50 & 87.50 & 32.67 & 47.57 & 67.31 & 62.73 & 55.04 \\
    Llama-3.3-70B-Instruct & 55.06 & 89.69 & 68.24 & 88.76 & 52.67 & 66.11 & 71.91 & 71.18 & 67.17 \\
    Qwen-2.5-7B-Instruct & 61.19 & 69.07 & 64.89 & 78.18 & 71.67 & 74.78 & 68.69 & 70.37 & 69.84 \\
    Qwen-2.5-14B-Instruct & 59.73 & 68.04 & 63.61 & 77.29 & 70.33 & 73.65 & 68.51 & 69.19 & 68.63 \\
    Qwen-2.5-32B-Instruct & 58.17 & 78.87 & 66.96 & 82.25 & 63.33 & 71.56 & 70.21 & 71.10 & 69.26 \\
    Qwen-2.5-72B-Instruct & 59.29 & 77.32 & 67.11 & 81.74 & 65.67 & 72.83 & 70.52 & 71.49 & 69.97 \\
    DeepSeek-R1-Distill-Llama-70B & 52.61 & 67.53 & 59.14 & 74.29 & 60.67 & 66.79 & 63.45 & 64.10 & 62.97 \\
    DeepSeek-R1 & 47.25 & 88.66 & 61.65 & 83.08 & 36.00 & 50.23 & 65.16 & 62.33 & 55.94 \\
    QwQ-32B & 49.70 & 86.08 & 63.02 & 82.91 & 43.67 & 57.21 & 66.31 & 64.87 & 60.11 \\
    Gemini-2.0-thinking & 59.52 & 77.32 & 67.26 & 81.82 & 66.00 & 73.06 & 70.67 & 71.66 & 70.16 \\
    o3-mini & 60.38 & 80.93 & 69.16 & 84.19 & 65.67 & 73.78 & 72.29 & 73.30 & 71.47 \\
    \hline
    \multicolumn{10}{c}{\textbf{French}} \\
    \hline
    GPT-3.5-turbo & 58.29 & 59.79 & 59.03 & 73.56 & 72.33 & 72.94 & 65.93 & 66.06 & 65.99 \\
    GPT-4-turbo & 68.75 & 51.03 & 58.58 & 72.86 & 85.00 & 78.46 & 70.80 & 68.02 & 68.52 \\
    GPT-4o & 63.59 & 67.53 & 65.50 & 78.12 & 75.00 & 76.53 & 70.86 & 71.26 & 71.02 \\
    GPT-4o-mini & 62.79 & 69.59 & 66.01 & 78.85 & 73.33 & 75.99 & 70.82 & 71.46 & 71.00 \\
    Claude-3.5-sonnet & 55.12 & 80.41 & 65.41 & 81.99 & 57.67 & 67.71 & 68.56 & 69.04 & 66.56 \\
    Claude-3.7-sonnet & 55.02 & 81.96 & 65.84 & 82.93 & 56.67 & 67.33 & 68.97 & 69.31 & 66.58 \\
    Gemini-1.5-pro & 70.95 & 54.12 & 61.40 & 74.28 & 85.67 & \textbf{79.57} & 72.61 & 69.90 & 70.49 \\
    DeepSeek-V3 & 63.35 & 62.37 & 62.86 & 75.91 & 76.67 & 76.29 & 69.63 & 69.52 & 69.57 \\
    Doubao-pro & 70.20 & 54.64 & 61.45 & 74.34 & 85.00 & 79.32 & 72.27 & 69.82 & 70.38 \\
    GLM-4-9B-chat & 53.96 & 38.66 & 45.05 & 66.48 & 78.67 & 72.06 & 60.22 & 58.66 & 58.55 \\
    Llama-3.1-8B-Instruct & 69.40 & 47.94 & 56.71 & 71.94 & 86.33 & 78.48 & 70.67 & 67.14 & 67.60 \\
    Llama-3.1-Tulu-3-8B & 50.96 & 68.56 & 58.46 & 73.82 & 57.33 & 64.54 & 62.39 & 62.95 & 61.50 \\
    Llama-3.1-70B-Instruct & 64.06 & 63.40 & 63.73 & 76.49 & 77.00 & 76.74 & 70.28 & 70.20 & 70.24 \\
    Llama-3.1-Nemotron-70B & 51.04 & 88.14 & 64.65 & 85.53 & 45.53 & 59.26 & 68.29 & 66.74 & 61.95 \\
    Llama-3.1-Tulu-3-70B & 49.69 & 83.51 & 62.31 & 80.95 & 45.33 & 58.12 & 65.32 & 64.42 & 60.21 \\
    Llama-3.3-70B-Instruct & 55.63 & 81.44 & 66.11 & 82.86 & 58.00 & 68.24 & 69.25 & 69.72 & 67.17 \\
    Qwen-2.5-7B-Instruct & 60.99 & 57.22 & 59.04 & 73.40 & 76.33 & 74.84 & 67.19 & 66.77 & 66.94 \\
    Qwen-2.5-14B-Instruct & 64.40 & 63.40 & 63.90 & 76.57 & 77.33 & 76.95 & 70.48 & 70.37 & 70.42 \\
    Qwen-2.5-32B-Instruct & 64.44 & 59.79 & 62.03 & 75.16 & 78.67 & 76.87 & 69.80 & 69.23 & 69.45 \\
    Qwen-2.5-72B-Instruct & 62.94 & 63.92 & 63.43 & 76.43 & 75.67 & 76.05 & 69.69 & 69.79 & 69.74 \\
    DeepSeek-R1-Distill-Llama-70B & 54.61 & 82.47 & 65.71 & 83.08 & 55.67 & 66.67 & 68.85 & 69.07 & 66.19 \\
    DeepSeek-R1 & 54.07 & 85.57 & 66.27 & 85.03 & 53.00 & 65.30 & 69.55 & 69.28 & 65.78 \\
    QwQ-32B & 59.77 & 78.87 & 68.00 & 82.77 & 65.67 & 73.23 & 71.27 & 72.27 & 70.62 \\
    Gemini-2.0-thinking & 63.52 & 76.29 & \textbf{69.32} & 82.38 & 71.67 & 76.65 & 72.95 & 73.98 & \textbf{72.98} \\
    o3-mini & 64.55 & 73.20 & 68.60 & 81.02 & 74.00 & 77.35 & 72.78 & 73.60 & \textbf{72.98} \\
    \hline
    \end{tabular}
  \end{adjustbox}
  \caption{The overall performance of representative LLMs in the English and French sections}
  \label{en french performance}
\end{table*}

\begin{table*}[htbp]
  \centering
  \begin{adjustbox}{max width=\textwidth}
\begin{tabular}{c|ccc|ccc|ccc}
    \hline
    \multirow{2}{*}{\textbf{Models}} & \multicolumn{3}{c|}{\textbf{RP}} & \multicolumn{3}{c|}{\textbf{IS}} & \multicolumn{3}{c}{\textbf{CE}} \\
     & P (\%) & R (\%) & F1 (\%) & P (\%) & R (\%) & F1 (\%) & P (\%) & R (\%) & F1 (\%) \\
     \hline
    \multicolumn{10}{c}{\textbf{German}} \\
    \hline
    GPT-3.5-turbo & 60.26 & 48.45 & 53.71 & 70.41 & 79.33 & 74.61 & 65.34 & 63.89 & 64.16 \\
    GPT-4-turbo & 67.14 & 48.45 & 56.29 & 71.75 & 84.67 & 77.68 & 69.45 & 66.56 & 66.98 \\
    GPT-4o & 64.06 & 63.40 & 63.73 & 76.49 & 77.00 & 76.74 & 70.28 & 70.20 & 70.24 \\
    GPT-4o-mini & 60.00 & 71.13 & 65.09 & 78.79 & 69.33 & 73.76 & 69.39 & 70.23 & 69.43 \\
    Claude-3.5-sonnet & 57.25 & 77.32 & 65.79 & 81.03 & 62.67 & 70.68 & 69.14 & 69.99 & 68.23 \\
    Claude-3.7-sonnet & 58.08 & 87.11 & 69.69 & 87.68 & 59.33 & 70.78 & 72.88 & 73.22 & 70.23 \\
    Gemini-1.5-pro & 64.10 & 51.55 & 57.14 & 72.19 & 81.33 & 76.49 & 68.15 & 66.44 & 66.82 \\
    DeepSeek-V3 & 62.63 & 61.34 & 61.98 & 75.33 & 76.33 & 75.83 & 68.98 & 68.84 & 68.90 \\
    Doubao-pro & 70.34 & 52.58 & 60.18 & 73.64 & 85.67 & \textbf{79.20} & 71.99 & 69.12 & 69.69 \\
    GLM-4-9B-chat & 53.96 & 56.19 & 55.05 & 70.89 & 69.00 & 69.93 & 62.43 & 62.59 & 62.49 \\
    Llama-3.1-8B-Instruct & 59.46 & 45.36 & 51.46 & 69.36 & 80.00 & 74.30 & 64.41 & 62.68 & 62.88 \\
    Llama-3.1-Tulu-3-8B & 50.34 & 75.77 & 60.49 & 76.73 & 51.67 & 61.75 & 63.54 & 63.72 & 61.12 \\
    Llama-3.1-70B-Instruct & 59.02 & 62.37 & 60.65 & 74.74 & 72.00 & 73.34 & 66.88 & 67.19 & 67.00 \\
    Llama-3.1-Nemotron-70B & 51.37 & 87.11 & 64.63 & 84.85 & 46.67 & 60.22 & 68.11 & 66.89 & 62.42 \\
    Llama-3.1-Tulu-3-70B & 50.16 & 82.99 & 62.52 & 80.92 & 46.67 & 59.20 & 65.54 & 64.83 & 60.86 \\
    Llama-3.3-70B-Instruct & 58.58 & 72.16 & 64.67 & 78.82 & 67.00 & 72.43 & 68.70 & 69.58 & 68.55 \\
    Qwen-2.5-7B-Instruct & 56.60 & 68.5 & 62.00 & 76.45 & 66.00 & 70.84 & 66.52 & 67.28 & 66.42 \\
    Qwen-2.5-14B-Instruct & 61.24 & 65.98 & 63.52 & 76.84 & 73.00 & 74.87 & 69.04 & 69.49 & 69.20 \\
    Qwen-2.5-32B-Instruct & 64.15 & 70.10 & 67.00 & 79.43 & 74.67 & 76.98 & 71.79 & 72.38 & 71.99 \\
    Qwen-2.5-72B-Instruct & 65.35 & 68.04 & 66.67 & 78.77 & 76.67 & 77.70 & 72.06 & 72.35 & 72.18 \\
    DeepSeek-R1-Distill-Llama-70B & 52.82 & 67.53 & 59.28 & 74.39 & 61.00 & 67.03 & 63.61 & 64.26 & 63.15 \\
    DeepSeek-R1 & 56.84 & 83.51 & 67.64 & 84.69 & 59.00 & 69.55 & 70.77 & 71.25 & 68.59 \\
    QwQ-32B & 55.40 & 79.38 & 65.25 & 81.48 & 58.67 & 68.22 & 68.44 & 69.02 & 66.74 \\
    Gemini-2.0-thinking & 62.71 & 76.29 & 68.84 & 82.17 & 70.67 & 75.99 & 72.44 & 73.48 & 72.41 \\
    o3-mini & 64.78 & 76.80 & \textbf{70.28} & 82.95 & 73.00 & 77.66 & 73.87 & 74.90 & \textbf{73.97} \\
    \hline
    \multicolumn{10}{c}{\textbf{Chinese}} \\
    \hline
    GPT-3.5-turbo & 58.52 & 68.37 & 63.06 & 76.78 & 68.33 & 72.31 & 67.65 & 68.35 & 68.65 \\
    GPT-4-turbo & 73.74 & 67.35 & 70.40 & 79.81 & 84.33 & 82.01 & 76.78 & 75.84 & 76.20 \\
    GPT-4o & 77.66 & 74.49 & 76.04 & 83.77 & 86.00 & \textbf{84.87} & 80.71 & 80.24 & \textbf{80.46} \\
    GPT-4o-mini & 68.12 & 71.94 & 69.98 & 80.97 & 78.00 & 79.46 & 74.54 & 74.97 & 74.72 \\
    Claude-3.5-sonnet & 60.14 & 90.82 & 72.36 & 91.00 & 60.67 & 72.80 & 75.57 & 75.74 & 72.58 \\
    Claude-3.7-sonnet & 59.73 & 89.29 & 71.57 & 89.66 & 60.67 & 72.37 & 74.69 & 74.98 & 71.97 \\
    Gemini-1.5-pro & 69.14 & 57.14 & 62.57 & 74.85 & 83.33 & 78.86 & 71.99 & 70.24 & 70.72 \\
    DeepSeek-V3 & 74.44 & 68.37 & 71.28 & 80.38 & 84.67 & 82.47 & 77.41 & 76.52 & 76.87 \\
    Doubao-pro & 77.68 & 87.00 & \textbf{82.08} & 75.62 & 61.73 & 67.98 & 76.65 & 74.37 & 75.03 \\
    GLM-4-9B & 64.71 & 33.67 & 44.30 & 67.01 & 88.00 & 76.08 & 65.86 & 60.84 & 60.19 \\
    Llama-3.1-8B-Instruct & 64.32 & 65.31 & 64.81 & 77.10 & 76.33 & 76.72 & 70.71 & 70.82 & 70.76 \\
    Llama-3.1-Tulu-3-8B & 84.44 & 38.00 & 52.41 & 48.48 & 89.29 & 62.84 & 66.46 & 63.64 & 57.63 \\
    Llama-3.1-70B-Instruct & 65.02 & 73.98 & 69.21 & 81.32 & 74.00 & 77.49 & 73.17 & 73.99 & 73.35 \\
    Llama-3.1-Nemotron-70B & 49.18 & 92.35 & 64.18 & 88.28 & 37.67 & 52.80 & 68.73 & 65.01 & 58.49 \\
    Llama-3.1-Tulu-3-70B & 54.02 & 85.71 & 66.27 & 84.86 & 52.33 & 64.74 & 69.44 & 69.02 & 65.51 \\
    Llama-3.3-70B-Instruct & 64.29 & 82.65 & 72.32 & 86.07 & 70.00 & 77.21 & 75.18 & 76.33 & 74.76 \\
    Qwen-2.5-7B-Instruct & 72.73 & 65.31 & 68.82 & 78.75 & 84.00 & 81.29 & 75.74 & 74.65 & 75.05 \\
    Qwen-2.5-14B-Instruct & 70.44 & 72.96 & 71.68 & 81.91 & 80.00 & 80.94 & 76.18 & 76.48 & 76.31 \\
    Qwen-2.5-32B-Instruct & 69.95 & 76.02 & 72.86 & 83.39 & 78.67 & 80.96 & 76.67 & 77.34 & 76.91 \\
    Qwen-2.5-72B-Instruct & 69.72 & 77.55 & 73.43 & 84.17 & 78.00 & 80.97 & 76.95 & 77.78 & 77.20 \\
    DeepSeek-R1-Distill-Llama-70B & 58.69 & 77.55 & 66.81 & 81.43 & 64.33 & 71.88 & 70.06 & 70.94 & 69.35 \\
    DeepSeek-R1 & 57.24 & 86.73 & 68.97 & 86.93 & 57.67 & 69.34 & 72.09 & 72.20 & 69.15 \\
    QwQ-32B & 52.34 & 85.71 & 64.99 & 84.00 & 49.00 & 61.89 & 68.17 & 67.36 & 63.44 \\
    Gemini-2.0-thinking & 66.81 & 80.10 & 72.85 & 85.06 & 74.00 & 79.14 & 75.93 & 77.05 & 76.00 \\
    o3-mini & 68.44 & 78.57 & 73.16 & 84.50 & 76.33 & 80.21 & 76.47 & 77.45 & 76.68 \\
    \hline
  \end{tabular}
  \end{adjustbox}
  \caption{The overall performance of representative LLMs in the German and Chinese sections}
  \label{german cn performance}
\end{table*}

\end{document}